\def\year{2019}\relax
\documentclass[letterpaper]{article} %
\usepackage{aaai19}  %
\usepackage{times}  %
\usepackage{helvet}  %
\usepackage{courier}  %
\usepackage{url}  %
\usepackage{graphicx}  %
\usepackage{amsmath}
\usepackage{amssymb}
\usepackage{multirow}

\begin{document}
\title{$A^2$-Net: Molecular Structure Estimation from Cryo-EM Density Volumes}
\author{Kui Xu$^1$, Zhe Wang$^2$,  Jianping Shi$^2$, Hongsheng Li$^3$, Qiangfeng Cliff Zhang$^1$\\
$^1$Tsinghua University $^2$SenseTime Research $^3$The Chinese University of Hong Kong\\
xuk16@mails.tsinghua.edu.cn, \{wangzhe, shijianping\}@sensetime.com, hsli@ee.cuhk.edu.hk, qczhang@tsinghua.edu.cn
}

\maketitle

\begin{abstract}
Constructing of molecular structural models from Cryo-Electron Microscopy (Cryo-EM) density volumes is the critical last step of structure determination by Cryo-EM technologies. Methods have evolved from manual construction by structural biologists to perform 6D translation-rotation searching, which is extremely compute-intensive. In this paper, we propose a learning-based method and formulate this problem as a vision-inspired 3D detection and pose estimation task. We develop a deep learning framework for amino acid determination in a 3D Cryo-EM density volume. We also design a sequence-guided Monte Carlo Tree Search (MCTS) to thread over the candidate amino acids to form the molecular structure. This framework achieves 91\% coverage on our newly proposed dataset and takes only a few minutes for a typical structure with a thousand amino acids. Our method is hundreds of times faster and several times more accurate than existing automated solutions without any human intervention. 
\end{abstract}

\section{Introduction}
\begin{figure}[t]
\centering
\includegraphics[width=\linewidth]{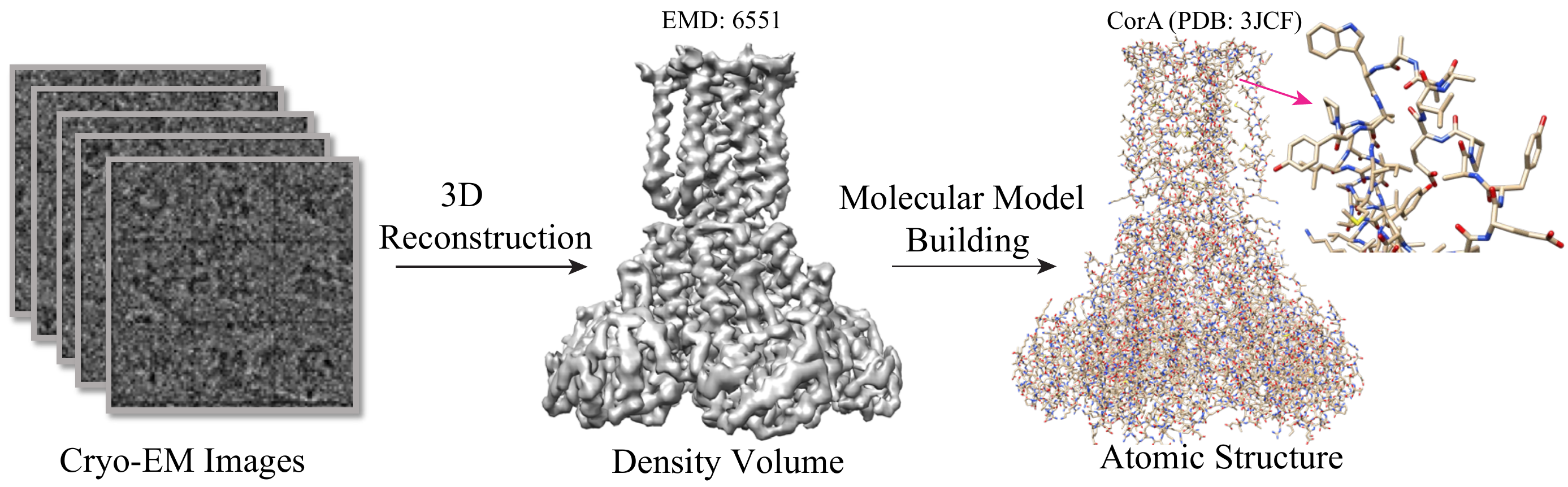}
\caption{The overview of resolving 3D atomic structures. }
\label{fig:overview}
\end{figure}

Resolving the 3D atomic structures of macro molecules is of fundamental importance to biological and medical research. Single particle Cryo-EM has emerged as a revolutionary technique that images biomolecules frozen in their native (or native-like) states. 
With Cryo-EM, 2D projection images are firstly collected and then reconstructed into a volumetric data, \emph{i.e.}, density volume, by software tools such as Relion \cite{cryoem-relion} and cryoSparc \cite{cryoem-cryospac-nmeth,cryoem-cryospac-tpami,cryoem-cryospac-cvpr}. Next, a molecular model that represents the atomic coordinates of each amino acid, the building blocks of protein molecules, is constructed and fitted into the 3D density volume (Fig. \ref{fig:overview}).

Despite the steady progresses towards automatic Cryo-EM structure determination, molecular model building remains a bottleneck. This step is difficult to automate since it relies substantially on human expertise. Structural biology experts are needed to manually assign specific amino acids to a density volume, with the help of 3D visualization tools, such as Chimera \cite{cryoem-chimera} and Coot \cite{cryoem-coot}. These manual operations are also time-consuming and inevitably error-prone. Attempts to automate this process include Rosetta \cite{rosetta-denovo} (we named as Rosetta-denovo), RosettaES \cite{cryoem-rosettaes}, Phinex \cite{cryoem-phenix} and EMAN2 \cite{cryoem-eman2}. However, their accuracy and coverage remain quite limited, often take hundreds of hours and frequently require human intervention. 

We approached this task from a novel perspective, inspired by the great success of deep learning applications in image recognition \cite{krizhevsky2012imagenet}, we choose to approach this task from a totally novel perspective. The molecular structure determination problem, in our view, can be considered as three sub-problems: 1) \textit{amino acid detection} in the density volumes, 2) \textit{atomic coordinates assignment} to determine the atomic coordinates of each amino acid and, 3) \textit{main chain threading} to resolve the sequential order of amino acids that form each protein chain. 

Leveraging the power of deep Convolutional Neural Network \cite{srivastava2015training}, we reformulated the problem and developed a novel framework for amino acid detection that learns the distribution of conformational densities of individual amino acids. 
Moreover, we designed a sequence-guided neighbor loss in training step to encode prior knowledge of protein sequences into amino acid detection. We also propose an MCTS algorithm to search and thread over the amino acids to form the full molecular structure. Our approach to molecular structure determination does not require human intervention and, when tested on a newly proposed large-scale dataset, it runs hundreds of times faster, and more accurately than existing methods. 
Finally, to the best of our knowledge, there remains no publicly available large-scale labeled dataset for the research of molecular structure determination from Cryo-EM density volumes. The dataset we collected and used in this study, named as the $A^2$ dataset, includes 250K amino acid objects in 1,713 protein chains from 218 structures. It constitutes a useful resource for evaluating molecular structure determination methods. 

To summarize, our contributions are four fold:
\begin{itemize}
\item This is the first attempt to formulate molecular structure determination from Cryo-EM density volumes with a deep learning approach.
\item We adapt a novel 3D network architecture for amino acid detection and internal atom coordinate estimation in density volumes, and proposed an APRoI layer and neighbor loss for better performance.
\item We design a sequence-guided MCTS algorithm for fast and accurate main chain threading. 
\item We will release a large scale, richly annotated dataset of protein density volumes, to facilitate research in this area.
\end{itemize}

\section{Related Work}
\subsubsection{Molecular Structure Determination} 
Structure determination of Cryo-EM maps is the process of generating a structure model with 3D coordinates for each atom in the macromolecule (e.g., proteins) that fits the map (Figure 1). The main approaches for molecular structure determination are $de$ $novo$ building and homology modeling, which its homologous structures in the Protein Database. In this work, we focus on the de novo approaches, where there are no previously solved structures of homologous proteins.

All recent molecular structure determination pipelines still rely on interactive tools with heavy hand labor. In principle, most available approaches, of which Rosetta-denovo is a typical example, use template matching and Monte Carlo sampling based on a library with millions of fragments from solved protein structures as templates for structural modeling. Briefly the target protein is divided into short fragments and structures of similar sequences in the library are identified for every fragment.
And then candidate structure fragments are assembled by Monte Carlo simulated annealing to optimize a fitting score.
Alternatively, RosettaES then uses a greedy conformational sampling algorithm to assemble the main chain of protein maximally consistent with sequence and density volume. The accuracy and coverage of these methods are often not satisfactory, due to limitations of the hand-crafted scoring functions and the sheer number of template structures.

\subsubsection{Object Detection} 
Approaches for object detection including Faster R-CNN, Cascade R-CNN, SNIPER, FishNet \cite{ren2015faster,cai18cascadercnn,sniper2018,fishnet} have improved drastically in terms of performance and efficiency. These methods follow a similar framework in which the objects are extracted from a Region of Interest (RoI) and pooled to the same size before predicting their categories and coordinates. RoI pooling, RoI Warping and RoIAlign~ \cite{girshick2014rich,dai2016instance,mask-rcnn} are popular techniques for RoI extraction, all of which break the original aspect ratio of the objects to account for their variations in natural images filmed with different angles and distances.
However, in some scenarios like in our work, aspect ratios of different types of amino acids should be preserved. To tackle this problem, we proposed an Aspect-Ratio Preserved RoI (APRoI) layer to capture the aspect ratio in amino acids.

3D object detection plays a key role in a variety of real-world applications, such as autonomous driving~\cite{gomez2016pl}, augmented/virtual reality and identification of disease diagnosis. MV3D \cite{cvpr17chen} focuses on very sparse data (LiDAR point cloud) and projects 3D data into 2D multi-view images. It is less effective in the amino acid detection task where difficulties may come from dense objects and the ambiguity in overlapping regions. VoxelNet \cite{zhou2017voxelnet} groups very sparse points for encoding voxel features to model point interactions. Frustum-PointNet \cite{qi2017frustum} extracts points within the frustum from 2D box to form a frustum point cloud, which may generate too much noise, especially in a dense object detection task. In this work, we designed a pure 3D detector.
\subsubsection{3D Pose Estimation} 
Given well labeled 3D joint locations, 3D pose estimation aims to determine the precise joint locations in 3D space. For instance, 3D human pose estimation attempts to regress 16 key points in the human body based on 3D joint locations of a human skeleton from 2D images \cite{pose_Zhou_2017_ICCV,pose_martinez_2017_3dbaseline}. However, the existing 3D pose estimation are based on 2D images. In this work, we introduce a pure 3D pose estimation module to produce atomic coordinates in volumetric data. 

Inspired by these breakthroughs, we designed a multi-task 3D neural network that first detects amino acids and then estimates the atom positions. We also took advantage of the sequence information and introduced a sequence-guided neighbor loss to train the network, which has not been explored before. In addition, we added a postprocessing with superior tree searching algorithm -- MCTS for main chain threading. We believe that this is the first attempt using MCTS to trace the boxes of amino acids guided by sequence. 

\section{Problem Definition}

Given a 3D Cryo-EM density volume and its protein sequence as inputs, our goal is to detect all the amino acids, estimate their pose and thread them into a protein chain in the 3D space. The density volume, obtained from 3D reconstruction of 2D microscope images, is represented as 3D matrices with continuous density value in each voxel. The sequential orders of amino acids, but not their locations, are known. Each amino acid can be represented by a class $C$ (20 types of standard amino acid), a 3D box and $N^C$ central locations of atoms in the amino acid. Each 3D box is parameterized by two coordinates: the front left top corner and the back right bottom corner. $N^C$ (4$\sim$14) is the number of the atoms in the amino acid, and it varies in different types.

\section{$A^2$ Dataset}

As a benchmark for molecular structure determination, we built a large-scale Amino Acid ($A^2$) dataset of Cryo-EM density volumes. It contains 250,000 amino acids in 1713 simulated (by Chimera in 3 \AA) electron density volumes and is annotated with rich information of amino acids. The amino acids are labeled with 3D boxes of 20 categories as well as the atomic coordinates. The amino acids in the dataset are dense, small objects that overlap each other, which makes the amino acid detection a very challenging task. To the best of our knowledge, the $A^2$ dataset is the first large-scale benchmark for learning automatic molecular structure determination.

\subsection{3D Density Volume Annotation}

The molecular structures and the corresponding density volumes in the $A^2$ dataset were collected from the RCSB Protein DataBase (PDB) and The Electron Microscopy Data Bank (EMDB). Firstly, we selected the volumes with resolution below 5 angstroms (\AA). The PDB and EMDB databases contain some inconsistencies where some volumes do not match the ground-truth structure. We manually removed these problematic volumes. As a quality control step, we collected only the chains without any missing atom or amino acid.
Ultimately, the $A^2$ dataset contained 250,000 amino acids in 1713 chains are kept to construct the $A^2$ dataset. Following random selection, we obtained a split of 1250 training and 463 validation chains.

\subsection{Dataset Statistics}

\begin{figure}
\centering
\includegraphics[width=0.95\linewidth]{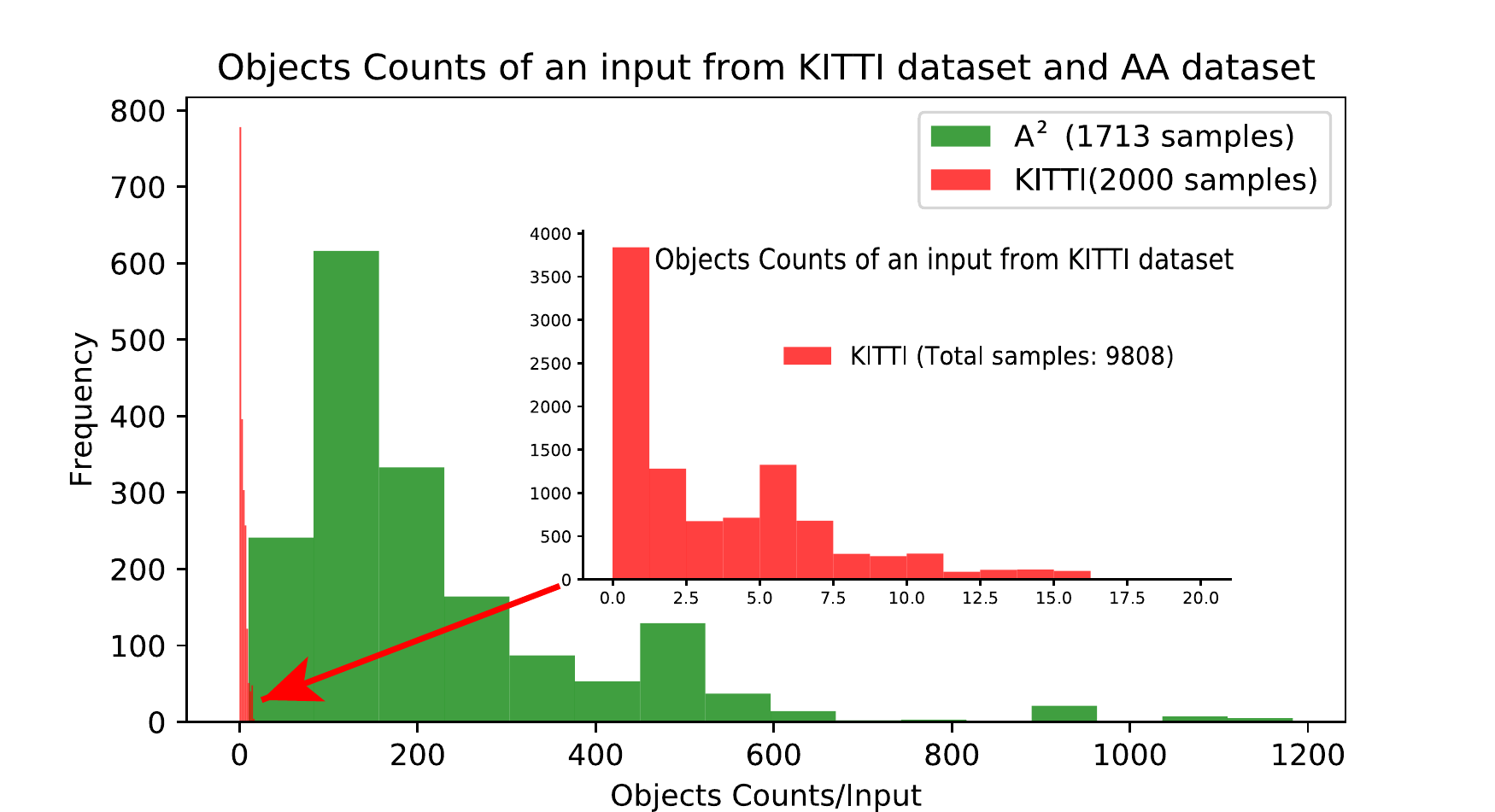}
\caption{The significant object density difference between the proposed dataset and KITTI LiDAR detection dataset.}
\label{fig:dense}
\end{figure}

There are 20 categories of common amino acids and 367,929 pairs of overlapping amino acids in the dataset, which means the dataset is highly dense and challenging for detection. As shown in Fig. \ref{fig:dense}, the dataset has much denser 3D objects than the KITTI dataset \cite{Geiger2012CVPR}.

\section{Method}
\label{sec:blind}
The framework of $A^2$-Net consists of two stages. Stage one represents the deep neural network for amino acid detection in 3D space and pose estimation, which determines the 3D coordinates of atoms in each amino acid. Stage two uses a Monte Carlo Tree Search strategy with tree pruning, based on the candidate amino acid proposals obtained in stage one to construct the main chains of amino acids, i.e., proteins.
\begin{figure*}[t]
\centering
\includegraphics[width=\linewidth]{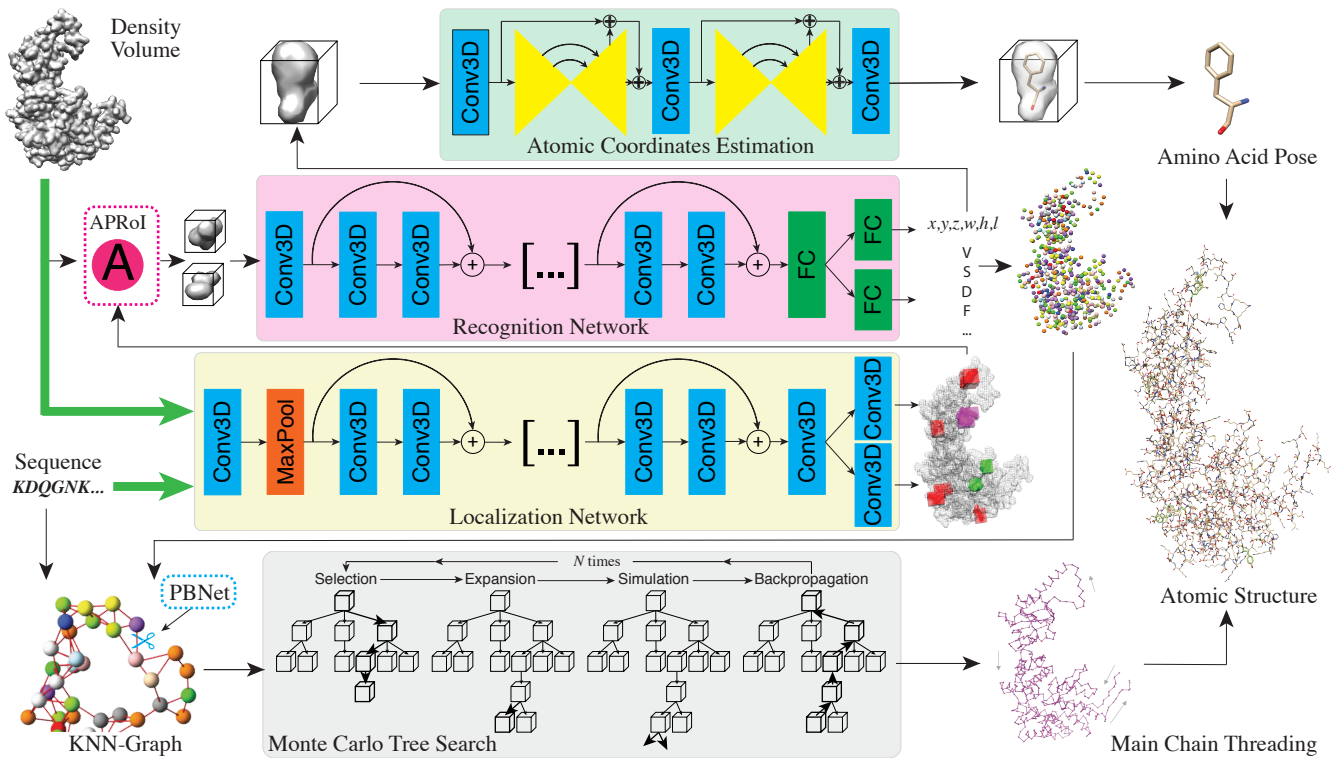}
\caption{The architecture of $A^2$-Net. Given the density volume and the amino acid sequential orders, the localization network (locNet) and recognition network (recNet)  locate and classify 20 types of amino acids in it. The atomic coordinates estimation network regresses the atomic coordinates. With the proposed amino acid, a MCTS algorithm is used for main chain threading.}
\label{fig:pipeline}
\end{figure*}

\subsection{3D Amino Acid Detection} \label{sec:net}
As shown in Fig. \ref{fig:pipeline}, when given a density volume, our $A^2$-Net first obtains 3D feature volumes and generates 3D box proposals with the region proposal network (RPN) \cite{ren2015faster}. The 3D RPN consists of three 3D convolutional layers to generate proposals of amino acid locations. We used 3D anchors at each 3D location to cover the region of amino acids with various scales and aspect ratios. Next, one branch of RPN classified whether the anchors are valid amino acid proposals and the other branch estimates their coordinates. With the amino acid proposals, we used a newly designed Aspect-ratio Preserved RoI (\textit{APRoI}) layer to extract the RoI in the input volume into a fixed cubic. Then the cubic went through several 3D convolutional layers to finally predict its amino acid category and coordinates. 

\subsubsection{Aspect-ratio Preserved RoI Layer}

In natural images, objects generally have completely different aspect ratios and conventional RoI pooling resizes the regions and abandons the original aspect ratio of the object. It can be seen as aspect-ratio augmentation, which is usually beneficial for the generalization ability of the deep model. However, aspect ratios of different amino acids should be maintained as they actually reflect different categories. We thus proposed an APRoI layer, which first crops the input at the RoI location, and then pads it with zero to a defined size of $W_T \times H_T \times L_T$. Despite its simplicity, the APRoI layer preserved the aspect ratio of the objects and proved to be vital for our amino acid classification.

The back-propagation passes derivatives through the APRoI layer. Let $x_{ijk} \in \mathbb{R}^3$ be the input of the APRoI layer, and $y_{rijk}$ be the layer's output at voxel index $i,j,k$ from the $r$-th RoI. 
The APRoI layers back-propagate partial derivative of the loss with respect to each input $x_{ijk}$ as Eqn \ref{eqn:aproi2}: 
\begin{align}
\begin{split}
   & \frac{\partial L}{\partial x_{ijk}} = \sum_r [i_s\le i < i_e, j_s\le j < j_e, k_s\le k < k_e]\frac{\partial L}{\partial y_{rijk}} ,\\
&i_s =max \lbrace {0, \lfloor\frac{W_T-W_r}{2} \rfloor} \rbrace, i_e =i_s + min \lbrace  \lceil W_r\rceil, W_T \rbrace ,\\
&j_s =max \lbrace {0, \lfloor\frac{H_T-H_r}{2} \rfloor} \rbrace, j_e =j_s + min \lbrace  \lceil H_r\rceil, H_T \rbrace,\\
&k_s =max \lbrace {0, \lfloor\frac{L_T-L_r}{2} \rfloor} \rbrace, k_e =k_s + min \lbrace  \lceil L_r\rceil, L_T \rbrace,
\end{split}\label{eqn:aproi2}
\end{align}
where $W_r, H_r,L_r$ are the size of the $r$-th RoI.  $\partial L / \partial y_{rijk}$ denotes the partial derivative computed by the layer on top of the APRoI layer. Intuitively, only the previously valid locations will receive the gradients for back propagation, while the locations with padded zeros are simply ignored.

\subsubsection{Loss for amino acid detection}
To predict the 3D box size, we follow previous work \cite{qi2017frustum} and use a mixture of both regression and classification formulations instead of directly regressing the 3D box size. Firstly,  we specifically pre-define 20 template size with ($w^c, h^c, l^c$). Our model classifies each input into one type and then regresses the residual of the box sizes of the type with highest probability. While for box center regression, we parametrize a 3D ground truth box as $(x^g, y^g, z^g, w^g, h^g, l^g)$, where $(x^g, y^g, z^g)$ represents the coordinates of the front left top corner of the box and $(w^g, h^g, l^g)$ represents width, height and length. We define the residual vector as $\textbf{u} \in \mathbb{R}^3$, which contains the 3 regression targets corresponding to center location $\Delta x, \Delta y$ and $\Delta z$, and define the residual vector as $\textbf{v} \in \mathbb{R}^3$ containing 3 dimensions ($\Delta w, \Delta h$, $\Delta l$). The residuals are computed as: $\Delta x = (x^g - x^a)/w_a, \Delta y = (y^g - y^a)/h_a$, $\Delta z =(z^g - z^a)/l_a, \Delta w = \log ((w^g-w^c)/w_a)$,  $\Delta h = \log ((h^g-h^c)/h_a), \Delta l =\log ((l^g-l^c)/l_a)$.

The proposed loss function is defined as:
\begin{align}
\begin{split}
    Loss &= \frac{1}{N_{cls}} \sum_i^{N_{cls}}  L_{neighbor}(p_i, p_i^*) +\\ 
     \beta  \frac{1}{N_{reg}} \sum_j^{N_{reg}} &p_j^* (L_{c-reg}(\textbf{u}_j, \textbf{u}_j^*)+ L_{s-reg}(\textbf{v}_j, \textbf{v}_j^*)),
\end{split}
\label{eqn:2}
\end{align}
where $p_i, p_i^*$ are respectively the predicted probability of anchor $i$ and its label, $u_j, u_j^*$ are the box center regression output and ground truth for anchor $j$, $v_j, v_j^*$ are the box size residual regression output and ground truth. $L_{neighbor}$ is the classification loss  reweighted by the guiding sequence, which will be introduced in the next section. $L_{c-reg}$, $L_{s-reg}$ are the center coordinates and residual box size regression loss (smooth $L_1$ loss), respectively. The two terms are normalized by $N_{cls}$ and $N_{res}$. $\beta$ is the balancing parameter.

\subsubsection{Neighbor Loss}\label{sec:neighborloss}
 
Since the sequential orders of the amino acids is provided in this task, the geometric constraints (distance, the overlapping region and sequential order) between amino acids should be integrated to regularize the detection results, as shown in \cite{gao2018question}.
In most cases, the distance between an amino acid and one of its two neighbors, should be smaller than the distance between its two neighbors. To take advantage of this information, we introduced a novel neighbor loss. For each proposal, we checked both criteria: 1) it is a positive anchor, and 2) its distance to either neighbor of the associated ground truth, is smaller than the distance between the two neighbors. The qualified anchor will be assigned a higher weight as they are ``better'' samples in the view of the sequence. The additional weight $(1-p_i)^\lambda$ follows the spirit of focal loss \cite{lin2017focal}, which down-weights the sample well classified by the model. We define the neighbor loss as:
\begin{align}
\begin{split}
      L_{neighbor}(p_i) = - ((1-p_i)^\lambda m_i + 1) log(p_i),
\end{split}    
\end{align}
where $m_i\in \{0,1\}$, and $m_i=1$ when object $i$ is one of the mined positive neighbor objects. 

\subsection{3D Amino Acid Pose Estimation}
After we obtained the proposal for an amino acid, we further estimated its pose by locating its forming atoms in 3D space. The Stacked hourglass Network~\cite{pose_hg} is widely used to handle the human pose estimation task. In this work, a 3D stacked hourglass network, named as poseNet, is proposed to regress 3D coordinates of each atom in amino acids. The network stacks multiple ($H$) hourglass structures sequentially. Each hourglass has $R_b$ residual blocks and provides feature volumes with different semantic resolutions. Importantly, auxiliary losses were applied to the intermediate feature volumes for learning robust features.

For an amino acid with $N$ atoms, poseNet produces $H$ estimated heatmaps with $N$ channels. The Mean Squared Error loss is adopted:
\begin{align}
\begin{split}
      L_{pose} = \sum_{h}^{H} \sum_{n}^{N} \parallel y_h^*(n) - y_h(n) \parallel_{2}^{2},
\end{split}    
\end{align}
where $y_h^*$ denotes the predicted heatmap by the $h$-th stack, $n$ denotes the $n$-th atom, and $y_h$ is the ground-truth heatmap with the $N \times 8 $ locations labeled as 1. For each atom, the $2^3$ neighborhood locations are labeled as 1.

\subsection{Monte Carlo Tree Search for Threading}\label{sec:threading}

\subsubsection{Main Chain Threading Problem } Given the predicted amino acid proposals and the ground truth sequential order, our next task is to select the same number of proposals as in the sequence and thread them over to form the complete protein chain. With $N_B$ predicted proposal set $B$ from the $A^2$-Net, and a sequence $S^*$ of length $T$ ($N_B > T$), our next task is to select $T$ boxes in sequence to form the complete protein chain. The categories and sequential orders of $S$ are given by $S^*$, and the proposals are selected from $B$. Each proposal is $S_t = (x_t, y_t, z_t, w_t, h_t, l_t, P_t)$,  $P_t$ is the probability of different categories predicted by $A^2$-Net.  

\subsubsection{Monte Carlo Tree Search} MCTS is a tree search algorithm in which a node is evaluated by performing random actions from the decision space until an outcome can be determined \cite{alphago,ReinforceWalk}. Searching by MCTS is done by iteratively building a search tree where the nodes denotes different states, and the edges are the actions leading to one state from another. A node is recursively added to the tree during each iteration. Based on the reward of the new node, the reward values of all parent nodes are updated. A single iteration of the MCTS building process consists of four steps: 1) selection: a node to be expanded is selected; 2) expansion: the node is expanded by simulating the associated action; 3) simulation: the tracing is simulated following a random path until the terminal amino acid is reached; 4) back propagation: the result propagates back through the tree.

\subsubsection{Building KNN-Graph} Directly performing the MCTS algorithm to all the proposals may be time consuming, so we first built a graph based on $K$ nearest neighbors, where each node denotes a proposal and an edge connects two nodes if they are among the $K$ nearest neighbors, thus called KNN graph. Next, we determined the root node of the tree by finding $L$ proposals in $B$ which match the first $L$ amino acids in the sequence. Finally, we obtained several candidate fragments as the starting points. For each starting point, we ran the MCTS algorithm to obtain the optimal path.
The optimal path was obtained by a control policy $\pi$ to maximize the total reward $R$, which is the sum of all the values $V_t$ in every following step $t$. The reward function $R$ thus is written as:

\begin{align}
\begin{split}
    R &= \sum_{t=1}^{T} V_t, \\
    V_t = t * ( P_{detection}(&S_t)  + P_{compatible}(S_t, S_{t+1}) ),
\end{split}\label{eqn:reward}
\end{align}
where $V_t$ is the sum of the detection score $P_{detection}$ and the compatible score $P_{compatibility}$ at action $t$, and weighted by the time $t$ which encourages the searching path to be long. $P_{detection}$ and $P_{compatibility}$ ensure that the selected boxes are reliable and compatible. $P_{detection}(t)$ is the detection probability of $S_t$, and $P_{compatibility} = dist(S_t, S_{t+1})$, where $dist(\cdot)$ is the IoU between two boxes. An optimal policy $\pi$ outputs an optimal action sequence, which is defined as a path with maximum reward from the root to a leaf. We seek a path that maximizes the reward function in Eqn. (\ref{eqn:reward}).

After the root was created, Monte Carlo simulations selected
actions and followed the sequence $S$ to create a new node. After a number of simulations, the tree was well populated, and the optimal path was selected. Each depth of the tree is the time step, with root at $t=1$ and leaves at $t = T$.

\begin{figure*}[]
\centering
\includegraphics[height=7cm]{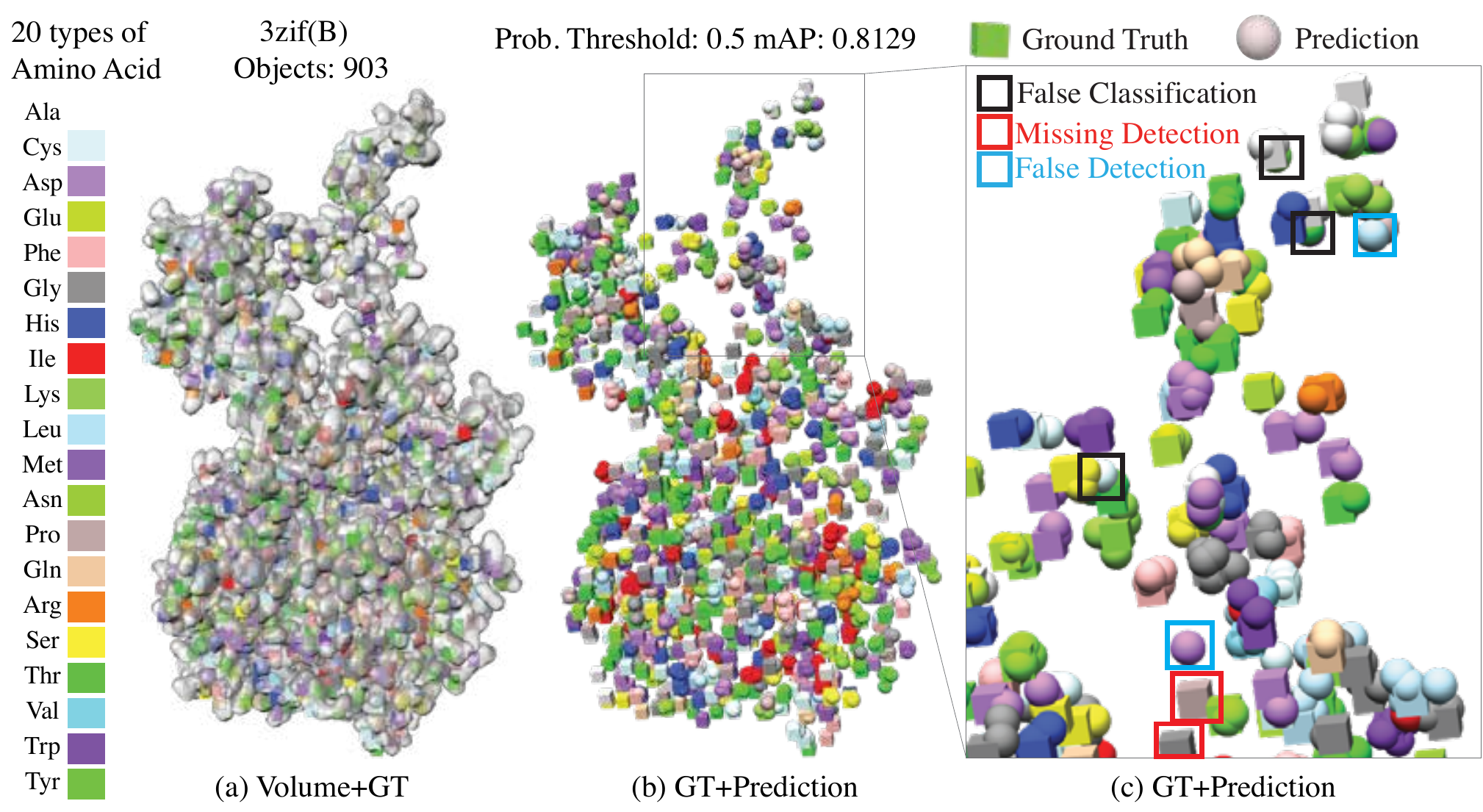}
\caption{ An example of amino acid detection result by $A^2$-Net. There are 903 amino acids in the volume (PDB id:3zif, B chain), the mAP is 0.8129 with the probability threshold 0.5. For better visualization, we use boxes with a fixed length to represent the center of the ground truth box and spheres with a fixed radius to represent the center of predicted box. 20 colors represent 20 types of amino acid. (a). Transparent density volume within the ground truth and prediction. (b). The ground truth and prediction on the whole volume. (c). The result of a local area with some bad cases labeled.  }
\label{fig:exp-detection}
\end{figure*}

\subsubsection{Next Action }
At time $t+1$, we need to select a proposal $S_{t+1}$ close enough to $S_t$ while keeping the category same as $S_{t+1}^*$.  Since there may be a few proposals for selection, we may face an exploration-exploitation dilemma, where the algorithm may fall into a local optimum. To balance this dilemma, we follow \cite{AIIDE1614003} to use the $Upper$ $Confidence$ $Bound$ $for$ $Trees$ (UCT) \cite{mcts} method to optimize the action selection.
 
The UCT method is designed for better action selection strategy. At step $t$, the next action $a_{t+1}$ is:
\begin{align}
  a_{t+1}  & = \arg \max_{a} ( \frac{V_a}{n_a} + C \sqrt{\frac{2 \ln N_a}{n_a}}),
\end{align}
where $V_{a}$ is the reward at action $a$. $N_a$ is the number of simulations that the node has been visited, and $n_a$ is the number of simulations that the node has been followed.

\subsubsection{Tree Pruning by Peptide Bond Recognition Network}
Since protein structure is highly twisted, a pair of amino acids that are close to each other in 3D space might be far apart in primary in the sequence. Experienced biologists distinguish whether two amino acids are connected or not, by examining whether there is a peptide bond between them. We also designed a Peptide Bond Recognition Network (PBNet) to predict whether there is a peptide bond between two proposals.  With PBNet, we can efficiently remove 50\% edges in the KNN graph on average,  which largely improves the search speed. PBNet has only three convolutional layers with batch normalization and max pooling, followed by three fully connected layers. The network is trained by $softmax$ loss. 

\section{Experiments}

\subsection{Implementation Details}

Our network architecture can be divided into three parts: localization network (locNet), recognition network (recNet) and pose estimation network (poseNet). 
In locNet, the backbone network is a fully convolutional network with 12 3D convolutional layers including 4 residual blocks. Since amino acids have a very small volume, we only have one max pooling layer.
We followed previous work \cite{ren2015faster} to design an anchor mechanism to cover various scales and aspect ratios of amino acids. We use 7 aspect ratios and 3 scales, yielding $k=21$ anchors at each position on the last $conv$ feature maps of the backbone network. We applied a $3\times 3\times 3$ convolutional layer to the $conv$ feature volumes, followed by two sibling $1\times 1\times 1$ convolution layers for classification and bounding box regression, respectively. Each anchor was assigned a binary label depending on whether it has an Intercession-over-Union (IoU) with a ground truth amino acid larger than a threshold 0.8.
The recNet also has 4 residual blocks, and three fully connected layers while the poseNet has 4-stacked hourglass. All the convolutional layers adopt the $3\times 3\times 3$ kernel size. The $A^2$-Net was trained in three stages. We first trained the locNet and poseNet individually for 100 epochs, and then fixed them while training the recNet for 400 epochs. Finally, we jointly optimize the whole $A^2$-Net with sequence-guided neighbor loss for another 400 epochs. We used Adam \cite{KingmaB14Adam} optimizer to train the model, starting by a learning rate of 0.0001, a momentum of 0.9 and a weight decay of 0.0001. We fine-tune our models with BatchNorm. We found that BatchNorm may reduce over-fitting. For each density volume, we randomly cropped a $64\times 64\times 64$ cube and send it into the network. Limited by the GPU memory, we set the batch size to be 1.

\subsection{Results on the $A^2$ dataset}

\subsubsection{Amino Acid Detection}

We first evaluated the effectiveness of APRoI layer and sequence-guided neighbor loss training. We adopted the commonly used mean Average Precision (mAP) for evaluation of detection.
The quantitative amino acid detection results are reported in Table. \ref{table:det}. 

\begin{table}[t]
\centering
\caption{The results of detection and threading comparing with other 3D object detection methods. }
\label{table:det}
\begin{tabular}{llc}
\hline\noalign{\smallskip}
Methods & mAP &  Coverage      \\
\noalign{\smallskip}
\hline
\noalign{\smallskip}
MV3D(BV+FV)   & 0.118 & 0.15\\
Frustum-Pointnet-v1   & 0.407 & 0.45\\
Frustum-Pointnet-v2  & 0.425 & 0.48\\\hline
3D-VGG+RoIpool8   & 0.360 & 0.32\\
3D-VGG+RoIpool8(w/o maxpool) & 0.423 & 0.41\\
3D-ResNet+RoIpool8   & 0.416 & 0.44\\
3D-ResNet+RoIpool8(Raw volume)   & 0.610 & 0.55\\
$A^2$-Net (APRoI8) & 0.711 & 0.67\\
$A^2$-Net w/o Neighbor Loss  & 0.865 & 0.72 \\
$A^2$-Net & \textbf{0.891} & \textbf{0.91}\\
\hline
\end{tabular}
\end{table}

\begin{table}[t]
\centering
\caption{The results of threading by DFS-based methods and the proposed MCTS+PBNet.}
\label{table:threading}
\begin{tabular}{llc}
\hline\noalign{\smallskip}
Methods &  Coverage       &  RMSD    \\
\noalign{\smallskip}
\hline
\noalign{\smallskip}
DFS$_o$  & 0.65 & 3.5 \\
DFS$_d$  & 0.68 & 3.1 \\
DFS$_d$+PBNet
   & 0.89 & 2.6 \\
MCTS  & 0.72 & 2.9 \\
MCTS+PBNet & \textbf{0.91} & \textbf{2.0 } \\\hline
\end{tabular}
\end{table}

We first directly applied 3D VGG with RoI pooling for amino acid detection, which only achieves 0.36 mAP, while 3D ResNet-10 \cite{he2015deep} achieves 0.416 mAP. 
The last $conv$ feature volumes were used for 3D region proposal. By analyzing the intermediate output, we found that the network seemed to be dominated by the 3D region proposal task, which made the last $conv$ feature volumes to be only sensitive to the existence of the amino acid. There was little category-specific information left in the feature volumes. So we directly performed RoI pooling on the raw input cube and trained the recNet, which achieved 0.610 mAP. This verified our assumption that the category-specific information may be discarded in the feature map of the locNet model. 

We then replaced the RoI pooling with APRoI layer, which further improved mAP to 0.711. This indicates the importance of preserving the aspect ratio for detection in this task. We also found that the output size of APRoI layer is important, as mAP improved when we changed the target size from $8^3$  to $16^3$, the mAP has a large improvement. Finally, we used the sequence-guided neighbor loss training strategy and further improved the mAP to 0.891. Fig. \ref{fig:exp-detection} shows a qualitative result of detection.  Although the gain of mAP from neighbor loss was only marginal, the sequence coverage percentage of the threading result improved substantially.  

\subsubsection{Main Chain Threading}
\begin{figure}
\centering
\includegraphics[width=0.9\linewidth]{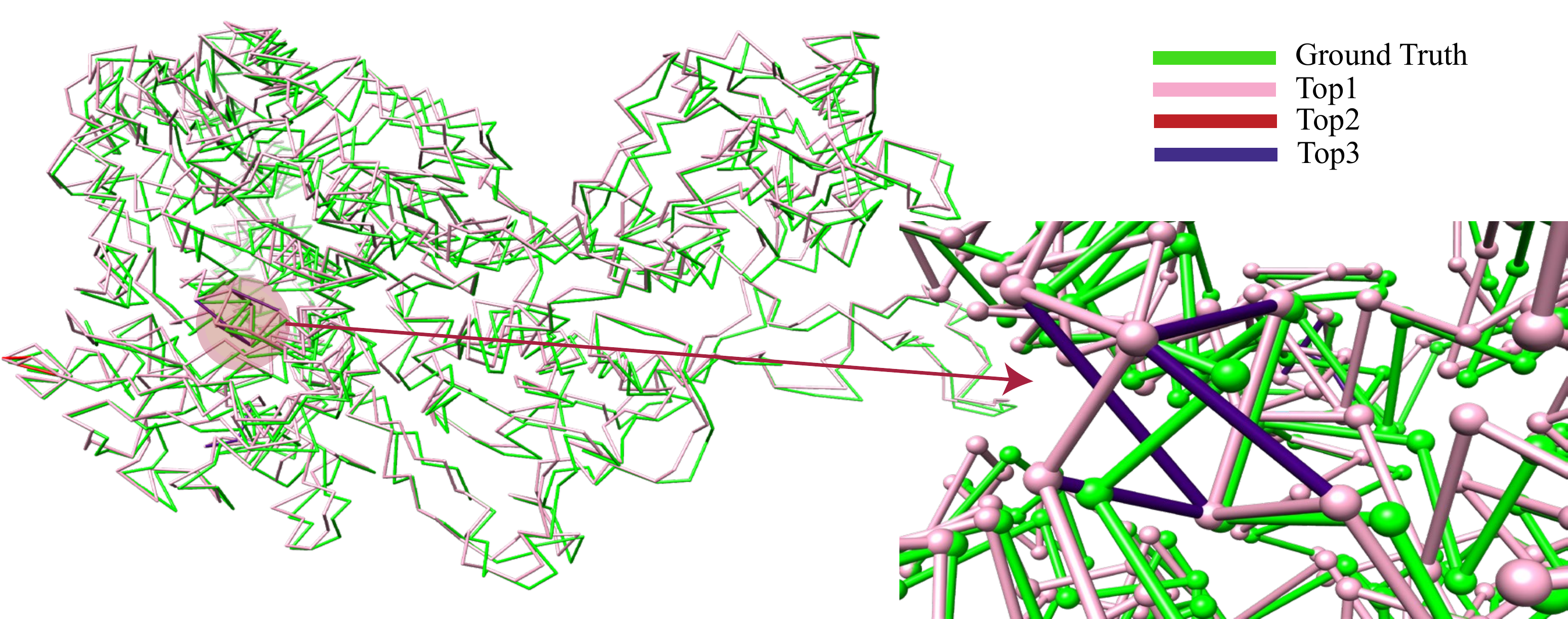}
\caption{An example of threading results by MCTS+PBNet.}
\label{fig:exp-threading}
\end{figure}

We mainly compared the proposed MCTS algorithm with a Depth First Search (DFS) method. DFS$_o$ and DFS$_d$ represent using IoU and distance as the selection criterion, respectively. The proposed MCTS algorithm outperformed the DFS based methods. The PBNet can be applied to both DFS and MCTS algorithms, as it is used to prune the trees by examining the existence of peptide bonds. It can be seen in Table. \ref{table:threading} shows that both threading algorithms are largely improved with PBNet. PBNet achieved 89.8\% accuracy for peptide bond recognition. In Fig. \ref{fig:exp-threading} shows some qualitative results of threading.

Table. \ref{table:Rosetta-denovo} demonstrates that Rosetta-denovo is very time-consuming. We ran Rosetta-denovo in a cluster with 200 computational nodes for 2 rounds. The CPU time was calculated by summing up all the tasks. It took Rosetta-denovo hundreds of hours to finish one round of computation, whereas our approach took only a few minutes and outperformed Rosetta-denovo by a huge margin.

\begin{table}[t]
\centering
\caption{Threading accuracy and efficiency compared with Rosetta-denovo. R1 and R2 denotes round 1 and 2. }
\label{table:Rosetta-denovo}
\begin{tabular}{l|cc|cc}
\hline
\multicolumn{1}{c|}{\multirow{2}{*}{Method}} & \multicolumn{2}{c|}{5gw5 (E) 528} & \multicolumn{2}{c}{4v19 (w) 166}    \\ \cline{2-5} 
\multicolumn{1}{c|}{}                        & Coverage       & Time             & Coverage & \multicolumn{1}{c}{Time} \\ \hline
Rosetta (R1)                              & 0.20           & 133 h       & 0.39     & 90 h               \\
Rosetta (R2)                              & 0.24           & 260 h       & 0.62     & 261 h             \\
Ours (MCTS)                               & \textbf{0.88}           & \textbf{11.3 m}       & \textbf{0.91}     & \textbf{6.8 m}                \\ \hline
\end{tabular}
\end{table}

\subsubsection{Comparison with other 3D Detection Methods} 
We adapted MV3D and Frustum-PointNet to the amino acid detection task. For MV3D, we cropped and projected the volumes into the bird's eye view and the front view (FV) and then trained MV3D. For Frustum-PointNet, we selected the voxels with density value is higher than the mean of the volumes as the points set. In the training step, we projected the 3D ground-truth boxes into their FV as the input. In the testing step, we projected the 3D boxes which were predicted by locNet into their FV as the input. Table. \ref{table:det} shows that our method outperformed them by a large margin. 

\subsubsection{Generic Features for 3D Detection} 

We pre-trained MV3D and Frustum-PointNet on the $A^2$ dataset, and fine-tuned them on KITTI dataset for 3D car bounding box regression. Table. \ref{table:improve_3ddet} summarizes the 3D car detection performance on the KITTI dataset. The $A^2$ dataset pre-trained model yielded an additional increase in performance, revealing that the $A^2$ dataset can provide generic features for 3D object detection task.

\begin{table}
\begin{center}
\caption{The AP of different methods for 3D car detection on KITTI dataset w/ or w/o $A^2$ Dataset. }
\label{table:improve_3ddet}
\begin{tabular}{lcccc}
\hline\noalign{\smallskip}
Methods & \begin{tabular}[c]{@{}c@{}} w/ $A^2$ \end{tabular} & Easy & Moderate & Hard \\
\noalign{\smallskip}
\hline
\noalign{\smallskip}

MV3D &   &  65.53 &58.97 & 59.14\\
MV3D &$\surd$  &\textbf{68.56} &\textbf{60.35} & \textbf{60.99}\\\hline
F-pointnet-v1 & & 83.26 & 69.28 & 62.56 \\
F-pointnet-v1 & $\surd$ &\textbf{84.89} & \textbf{71.97} & \textbf{64.07}\\\hline
F-pointnet-v2 & &  83.76 & 70.92 & 63.65 \\
F-pointnet-v2 & $\surd$ &\textbf{85.11} &\textbf{72.13} & \textbf{64.24} \\\hline
\end{tabular}
\end{center}
\end{table}

\section{Conclusions}
In this work, we reformulate the challenging molecular structure determination problem and propose a learning-based framework. The newly designed $A^2$-Net predicts accurate amino acid proposals with our APRoI layer and the neighbor loss training strategy. With the predictions and the sequence, we propose a MCTS algorithm for efficient threading. Using the peptide bond recognition network, tree branches between candidate pairs of proposals without a real peptide bond can be easily removed, which simultaneously improves the searching efficiency and the sequence coverage. Our novel method is hundreds of times faster and more accurate than the previous method, and will play a vital role in molecular structure determination.

\section{Acknowledgments}
This project is supported by the National Natural Science Foundation of China (Grants No. 31671355, 91740204, and 31761163007), the Beijing Advanced Innovation Center for Structural Biology, the Tsinghua-Peking Joint Center for Life Sciences and the National Thousand Young Talents Program. We thank Chuangye Yan, Xingyu Zeng, Yao Xiao and Xinge Zhu for their helpful work and insightful discussions.

\bibliographystyle{aaai19-xk}
\bibliography{aaai19-xk}

\begin{thebibliography}{}

\bibitem[\protect\citeauthoryear{Adams \bgroup et al\mbox.\egroup
  }{2010}]{cryoem-phenix}
Adams, P.~D.; Afonine, P.~V.; Bunk{\'{o}}czi, G.; and et~al.
\newblock 2010.
\newblock {{\it PHENIX}: a comprehensive Python-based system for macromolecular
  structure solution}.
\newblock {\em Acta Crystallographica Section D} 66(2):213--221.

\bibitem[\protect\citeauthoryear{Brubaker, Punjani, and
  Fleet}{2015}]{cryoem-cryospac-cvpr}
Brubaker, M.~A.; Punjani, A.; and Fleet, D.~J.
\newblock 2015.
\newblock Building proteins in a day: Efficient 3d molecular reconstruction.
\newblock In {\em CVPR}.

\bibitem[\protect\citeauthoryear{Cai and Vasconcelos}{2018}]{cai18cascadercnn}
Cai, Z., and Vasconcelos, N.
\newblock 2018.
\newblock Cascade r-cnn: Delving into high quality object detection.
\newblock In {\em CVPR}.

\bibitem[\protect\citeauthoryear{Chen \bgroup et al\mbox.\egroup
  }{2017}]{cvpr17chen}
Chen, X.; Ma, H.; Wan, J.; Li, B.; and Xia, T.
\newblock 2017.
\newblock Multi-view 3d object detection network for autonomous driving.
\newblock In {\em CVPR}.

\bibitem[\protect\citeauthoryear{Dai, He, and Sun}{2016}]{dai2016instance}
Dai, J.; He, K.; and Sun, J.
\newblock 2016.
\newblock Instance-aware semantic segmentation via multi-task network cascades.
\newblock In {\em CVPR}.

\bibitem[\protect\citeauthoryear{Devlin \bgroup et al\mbox.\egroup
  }{2016}]{AIIDE1614003}
Devlin, S.; Anspoka, A.; Sephton, N.; Cowling, P.; and Rollason, J.
\newblock 2016.
\newblock Combining gameplay data with monte carlo tree search to emulate human
  play.
\newblock In {\em AAAI}.

\bibitem[\protect\citeauthoryear{Emsley \bgroup et al\mbox.\egroup
  }{2010}]{cryoem-coot}
Emsley, P.; Lohkamp, B.; Scott, W.~G.; and Cowtan, K.
\newblock 2010.
\newblock Features and development of coot.
\newblock {\em Acta Crystallographica Section D - Biological Crystallography}
  66.

\bibitem[\protect\citeauthoryear{Frenz \bgroup et al\mbox.\egroup
  }{2017}]{cryoem-rosettaes}
Frenz, B.; Walls, A.~C.; Egelman, E.~H.; Veesler, D.; and DiMaio, F.
\newblock 2017.
\newblock Rosettaes: a sampling strategy enabling automated interpretation of
  difficult cryo-em maps.
\newblock {\em Nature Methods} 14:797.

\bibitem[\protect\citeauthoryear{Gao \bgroup et al\mbox.\egroup
  }{2018}]{gao2018question}
Gao, P.; Li, H.; Li, S.; Lu, P.; Li, Y.; Hoi, S.~C.; and Wang, X.
\newblock 2018.
\newblock Question-guided hybrid convolution for visual question answering.
\newblock In {\em ECCV}.

\bibitem[\protect\citeauthoryear{Geiger, Lenz, and
  Urtasun}{2012}]{Geiger2012CVPR}
Geiger, A.; Lenz, P.; and Urtasun, R.
\newblock 2012.
\newblock Are we ready for autonomous driving? the kitti vision benchmark
  suite.
\newblock In {\em CVPR}.

\bibitem[\protect\citeauthoryear{Girshick \bgroup et al\mbox.\egroup
  }{2014}]{girshick2014rich}
Girshick, R.; Donahue, J.; Darrell, T.; and Malik, J.
\newblock 2014.
\newblock Rich feature hierarchies for accurate object detection and semantic
  segmentation.
\newblock In {\em CVPR}.

\bibitem[\protect\citeauthoryear{Gomez-Ojeda, Briales, and
  Gonzalez-Jimenez}{2016}]{gomez2016pl}
Gomez-Ojeda, R.; Briales, J.; and Gonzalez-Jimenez, J.
\newblock 2016.
\newblock Pl-svo: Semi-direct monocular visual odometry by combining points and
  line segments.
\newblock In {\em IROS}.

\bibitem[\protect\citeauthoryear{He \bgroup et al\mbox.\egroup
  }{2016}]{he2015deep}
He, K.; Zhang, X.; Ren, S.; and Sun, J.
\newblock 2016.
\newblock Deep residual learning for image recognition.
\newblock In {\em CVPR}.

\bibitem[\protect\citeauthoryear{He \bgroup et al\mbox.\egroup
  }{2017}]{mask-rcnn}
He, K.; Gkioxari, G.; Doll{\'{a}}r, P.; and Girshick, R.~B.
\newblock 2017.
\newblock Mask {R-CNN}.
\newblock In {\em ICCV}.

\bibitem[\protect\citeauthoryear{Kingma and Ba}{2015}]{KingmaB14Adam}
Kingma, D.~P., and Ba, J.
\newblock 2015.
\newblock Adam: {A} method for stochastic optimization.
\newblock In {\em ICLR}.

\bibitem[\protect\citeauthoryear{Kocsis and Szepesv{\'a}ri}{2006}]{mcts}
Kocsis, L., and Szepesv{\'a}ri, C.
\newblock 2006.
\newblock Bandit based monte-carlo planning.
\newblock In {\em ECML}.

\bibitem[\protect\citeauthoryear{Krizhevsky, Sutskever, and
  Hinton}{2012}]{krizhevsky2012imagenet}
Krizhevsky, A.; Sutskever, I.; and Hinton, G.~E.
\newblock 2012.
\newblock Imagenet classification with deep convolutional neural networks.
\newblock In {\em NIPS}.

\bibitem[\protect\citeauthoryear{Lin \bgroup et al\mbox.\egroup
  }{2017}]{lin2017focal}
Lin, T.-Y.; Goyal, P.; Girshick, R.; He, K.; and Doll{\'a}r, P.
\newblock 2017.
\newblock Focal loss for dense object detection.
\newblock In {\em ICCV}.

\bibitem[\protect\citeauthoryear{Martinez \bgroup et al\mbox.\egroup
  }{2017}]{pose_martinez_2017_3dbaseline}
Martinez, J.; Hossain, R.; Romero, J.; and Little, J.~J.
\newblock 2017.
\newblock A simple yet effective baseline for 3d human pose estimation.
\newblock In {\em ICCV}.

\bibitem[\protect\citeauthoryear{Newell, Yang, and Deng}{2016}]{pose_hg}
Newell, A.; Yang, K.; and Deng, J.
\newblock 2016.
\newblock Stacked hourglass networks for human pose estimation.
\newblock In {\em ECCV}.

\bibitem[\protect\citeauthoryear{Pettersen \bgroup et al\mbox.\egroup
  }{2004}]{cryoem-chimera}
Pettersen, E.~F.; Goddard, T.~D.; Huang, C.~C.; and et~al.
\newblock 2004.
\newblock Ucsf chimera: A visualization system for exploratory research and
  analysis.
\newblock {\em Journal of Computational Chemistry} 25(13):1605--1612.

\bibitem[\protect\citeauthoryear{Punjani \bgroup et al\mbox.\egroup
  }{2017}]{cryoem-cryospac-nmeth}
Punjani, A.; Rubinstein, J.~L.; Fleet, D.~J.; and Brubaker, M.~A.
\newblock 2017.
\newblock cryosparc: algorithms for rapid unsupervised cryo-em structure
  determination.
\newblock {\em Nature Methods} 14:290.

\bibitem[\protect\citeauthoryear{Punjani, Brubaker, and
  Fleet}{2017}]{cryoem-cryospac-tpami}
Punjani, A.; Brubaker, M.~A.; and Fleet, D.~J.
\newblock 2017.
\newblock Building proteins in a day: Efficient 3d molecular structure
  estimation with electron cryomicroscopy.
\newblock {\em TPAMI} 39(4):706--718.

\bibitem[\protect\citeauthoryear{Qi \bgroup et al\mbox.\egroup
  }{2018}]{qi2017frustum}
Qi, C.~R.; Liu, W.; Wu, C.; Su, H.; and Guibas, L.~J.
\newblock 2018.
\newblock Frustum pointnets for 3d object detection from rgb-d data.
\newblock In {\em CVPR}.

\bibitem[\protect\citeauthoryear{Ren \bgroup et al\mbox.\egroup
  }{2015}]{ren2015faster}
Ren, S.; He, K.; Girshick, R.; and Sun, J.
\newblock 2015.
\newblock Faster r-cnn: Towards real-time object detection with region proposal
  networks.
\newblock In {\em NIPS}.

\bibitem[\protect\citeauthoryear{Scheres}{2012}]{cryoem-relion}
Scheres, S.~H.
\newblock 2012.
\newblock A bayesian view on cryo-em structure determination.
\newblock {\em Journal of Molecular Biology} 415(2):406 -- 418.

\bibitem[\protect\citeauthoryear{Shen \bgroup et al\mbox.\egroup
  }{2018}]{ReinforceWalk}
Shen, Y.; Chen, J.; Huang, P.-S.; Guo, Y.; and Gao, J.
\newblock 2018.
\newblock {ReinforceWalk: Learning to Walk in Graph with Monte Carlo Tree
  Search}.
\newblock In {\em ICLR}.

\bibitem[\protect\citeauthoryear{Shuyang~Sun}{2018}]{fishnet}
Shuyang~Sun, Jiangmiao~Pang, J. S. S. Y. W.~O.
\newblock 2018.
\newblock Fishnet: A versatile backbone for image, region, and pixel level
  prediction.
\newblock In {\em NIPS}.

\bibitem[\protect\citeauthoryear{Silver \bgroup et al\mbox.\egroup
  }{2016}]{alphago}
Silver, D.; Huang, A.; Maddison, C.~J.; Guez, A.; and et~al.
\newblock 2016.
\newblock Mastering the game of go with deep neural networks and tree search.
\newblock {\em Nature} 529:484.

\bibitem[\protect\citeauthoryear{Singh, Najibi, and Davis}{2018}]{sniper2018}
Singh, B.; Najibi, M.; and Davis, L.~S.
\newblock 2018.
\newblock {SNIPER}: Efficient multi-scale training.
\newblock In {\em NIPS}.

\bibitem[\protect\citeauthoryear{Srivastava, Greff, and
  Schmidhuber}{2015}]{srivastava2015training}
Srivastava, R.~K.; Greff, K.; and Schmidhuber, J.
\newblock 2015.
\newblock Training very deep networks.
\newblock In {\em NIPS}.

\bibitem[\protect\citeauthoryear{Tang \bgroup et al\mbox.\egroup
  }{2007}]{cryoem-eman2}
Tang, G.; Peng, L.; Baldwin, P.~R.; Mann, D.~S.; Jiang, W.; Rees, I.; and
  Ludtke, S.~J.
\newblock 2007.
\newblock Eman2: An extensible image processing suite for electron microscopy.
\newblock {\em Journal of Structural Biology} 157(1):38 -- 46.

\bibitem[\protect\citeauthoryear{Wang \bgroup et al\mbox.\egroup
  }{2015}]{rosetta-denovo}
Wang, R. Y.-R.; Kudryashev, M.; Li, X.; Egelman, E.~H.; Basler, M.; Cheng, Y.;
  Baker, D.; and DiMaio, F.
\newblock 2015.
\newblock De novo protein structure determination from near-atomic-resolution
  cryo-em maps.
\newblock {\em Nature Methods} 12:335--338.

\bibitem[\protect\citeauthoryear{Zhou and Tuzel}{2018}]{zhou2017voxelnet}
Zhou, Y., and Tuzel, O.
\newblock 2018.
\newblock Voxelnet: End-to-end learning for point cloud based 3d object
  detection.
\newblock In {\em CVPR}.

\bibitem[\protect\citeauthoryear{Zhou \bgroup et al\mbox.\egroup
  }{2017}]{pose_Zhou_2017_ICCV}
Zhou, X.; Huang, Q.; Sun, X.; Xue, X.; and Wei, Y.
\newblock 2017.
\newblock Towards 3d human pose estimation in the wild: A weakly-supervised
  approach.
\newblock In {\em ICCV}.

\end{thebibliography}

\end{document}